\documentclass[12pt]{article}

\usepackage{newtxtext,newtxmath}

\usepackage{graphicx}

\usepackage[letterpaper,margin=1in]{geometry}

\renewenvironment{abstract}
	{\quotation}
	{\endquotation}

\date{}

\makeatletter
\renewcommand{\fnum@figure}{\textbf{Figure \thefigure}}
\renewcommand{\fnum@table}{\textbf{Table \thetable}}
\makeatother

\usepackage{scicite}

\usepackage{url}

\newcommand{\FE}{\text{FE}}

\def\scititle{
Understanding generative AI content with embedding models
}

\title{\bfseries \boldmath \scititle}

\author{
	Max~Vargas$^{1\dagger}$,
	Reilly~Cannon$^{1\dagger}$,
	Andrew~Engel$^{1}$,
        Anand~D.~Sarwate$^{2\ast}$,
        Tony~Chiang$^{1, 3\ast}$\and
	\small$^{1}$Pacific Northwest National Laboratory, Richland, WA 99354, USA.\and
	\small$^{2}$Department of Electrical and Computer Engineering, Rutgers University, New Brunswick, NJ 08901, USA.\and
        \small$^{3}$ Department of Mathematics, University of Washington, Seattle, WA 98195, USA.\and
	\small$^\ast$Corresponding authors. Email: asarwate@rutgers.edu, chiang@math.washington.edu\and
	\small$^\dagger$These authors contributed equally to this work.
}

\begin{document} 

\maketitle

\begin{abstract} \bfseries \boldmath
Constructing high-quality features is critical to any quantitative data analysis. While \emph{feature engineering} was historically addressed by carefully hand-crafting data representations based on domain expertise, deep neural networks (DNNs) now offer a radically different approach. DNNs implicitly engineer features by transforming their input data into hidden feature vectors called embeddings. 
For embedding vectors produced by foundation models --- which are trained to be useful across many contexts --- we demonstrate that simple and well-studied dimensionality-reduction techniques such as Principal Component Analysis uncover inherent heterogeneity in input data concordant with human-understandable explanations. 
Of the many applications for this framework, we find empirical evidence that there is intrinsic separability between real samples and those generated by artificial intelligence (AI).
\end{abstract}

\noindent Modern deep neural networks (DNNs) have made striking progress in synthesizing image, language, and video data \cite{openai2024gpt4ocard, liu2024sorareviewbackgroundtechnology, stable-diffusion}. With larger datasets, DNNs continue to learn rich representations of the world characterized by their training data ~\cite{BengioRepLearning, singh2022flavafoundationallanguagevision, 8949228}. Their internal representations have been shown to be sensitive to high level concepts such as time and space~\cite{gurnee2024language}, color~\cite{li-etal-2021-implicit}, and more~\cite{li2023emergent, nanda-etal-2023-emergent}. Confidently relying on decisions made by DNNs requires establishing computationally accessible techniques to audit these generative outputs.  
In this paper, we leverage the numerical representations constructed by a foundation model's \emph{feature embedder} (FE), as shown in Figure 1A, to uncover semantic structure of our source data. Here, the input data to the FE is often text or images created from one or more other DNNs whose behavior we wish to understand. We show that dimensionality reduction techniques applied these embedding vectors reveal semantic distinctions in the underlying data, implying that the embedding space can be mined to perform tasks without additional training of the FE. For example, we find that embeddings from a foundation model's FE are naturally sensitive to whether input data originates from human sources or is generated by DNNs, showing that embeddings can potentially be used for deepfake detection without additional training.

Using linear model theory to explore the geometry of vectorized samples produced by these FEs, we find empirical evidence that 
the sampling techniques used to curate a dataset play a critical role in determining the resulting embedding vectors. In particular, datasets sampled from different sources often embed into separable subspaces of the ambient space. A similar phenomenon appears in other areas of science, such as a batch effect in experimental biology, where data collected from different trials of the same experiment exhibit significant variation. In natural language contexts, this sample diversity can be driven by intuitive concepts like subject matter or the appearances of specific words. Applying these techniques to study data produced by generative models, we observe fundamental differences from reference sets of real data. Furthermore, we demonstrate that the embeddings produced by FEs encode differences between generative techniques, such as changes to the generative model or prompt. These choices contribute to a kind of `generative DNA' (gDNA) of artificial data. By studying these differences through the numerical vector embeddings of FEs, we can understand the sources of these differences in terms of patterns in the observed data.


\subsection*{Understanding AI through feature embeddings}

DNNs process data through layers (stages) that 
successively map the input into a series of intermediate vector representations. The FE comprises the bulk of these operations, typically encompassing all but the final layer. After the FE recasts the data from its original format (text, images, etc.) into embedding vectors, a final task-specific ``head'' makes predictions using these embeddings~\cite{naveed2024comprehensiveoverviewlargelanguage, Belinkov2021ProbingCP, Chen2020GenerativePF}. 
The FEs corresponding to large language models (LLMs) or computer vision models may produce embedding vectors with thousands of dimensions. 


We use FEs from pre-trained models to show that the embedding vectors of different populations can be readily separated. Though we focus on text and image data, our techniques are applicable in many contexts where DNNs are used.
We also use FEs in two ways. First, a FE is part of a data generator. For LLMs, this is done by appending a next-token predictor to the FE. Second, a FE is a discriminating tool to analyze samples of data (Figure~1A). Using a discriminating FE, separability between the populations of real and AI-generated data means that AI-generated data can be identified using embedding vectors even when the FE is not fine-tuned to detect this difference.
That we can distinguish AI-generated data from real data has the implication that synthetic data does not faithfully reflect patterns found in real data. We show how unsupervised and supervised techniques can be used to this end. Further, applying techniques like an isolation forest~\cite{isolation-forest} to vectorized data, we show that outlier detection tools can be used to recognize AI-generated data.

\subsection*{Feature Embedders highlight shifted representations}
Pretrained FEs are sensitive to changes in sampling techniques that can go easily unnoticed by humans. Figure 1B shows that samples of cat images from two publicly available datasets are easily distinguishable when vectorized by a FE, in this case Apple's Data Filtering Network (DFN)~\cite{data-filtering-networks}. Despite both samples being comprised of real cat images, the vectors from the LSUN~\cite{yu2015lsun} dataset are starkly distinct from those created from the Cats\&Dogs~\cite{catsvsdogs} dataset. In fact, this separability is a dominant factor in the overall variability within the embedded data, with a linear separator easily identified by visually inspecting the first principal component (PC). However, more precise factors to explain this shift are hard to determine and may ultimately be a cumulative effect of differences in how the images were created (such as the number of pixels or the types of cameras used) and how they were processed and collated into each dataset. Some differences, such as those due to inconsistencies in data processing, might be subdued with enough knowledge to uniformly normalize the data. Other sources of variation are difficult to control, like effects of image watermarking and differences in camera hardware~\cite{StammWL:13forensics}. We note that different pre-trained FEs will likely vary in their ability to explicate these differences (Figure \ref{table:LDA_Embeddings_Comparison}).


When the dominant sources of variation are known, it can still be difficult to appropriately normalize differences between samples from different populations. 
Creating embedding vectors from Spanish and Russian news data, we initially see that the samples from these two languages are clearly separable by unsupervised clustering of the vectors (Figure 1C, c.f.~\ref{fig:other-langs}). This reflects the intuitive understanding that the two languages use different alphabets. We then used a machine translation model to remove the obvious distinction between languages and normalize the two kinds of data. Finding that the resulting Russian-to-Spanish translations are still separable from the native Spanish articles, we investigate further sources of variation. Topic analysis reveals that some of the remaining discrepancy between the native and translated Spanish text can be attributed to the content presented in each set of articles. For example, Russian economics articles discussed the Ruble while Spanish articles focused on the Euro --- a clear distinction that was not removed by the translation model (Figure \ref{fig:nmf_mlsum_factors}). Additional experiments hint that the embedding vectors detect stylistic differences between the native and translated articles: when using a FE to compare the translations produced by different language models, we are able to classify the different translations according to their source model (Table \ref{tab:lda_accs}). This observation  --- that FE can detect when different language models carry distinguishable linguistic patterns ---  suggests that none of the models perfectly match the patterns of human translators.




\subsection*{Tracing samples to sources with Generative DNA}
In the last section we observed that viewing data through a FE allows us to easily identify differences in samples collected from different sources, and that these differences persist even after attempts to standardize them. Using a discriminating FE to identify shifts between data sampled from different generative models, we find evidence of `generative DNA' (gDNA) within collected samples that allows us to easily distinguish real and synthetic data and even synthetic data created by different generative models. In particular, our observations on translation models are general phenomena. We tasked three LLMs --- Llama-2 70B~\cite{touvron2023llama2openfoundation}, Mixtral 8x7B~\cite{jiang2024mixtralexperts}, and Falcon 40B~\cite{almazrouei2023falconseriesopenlanguage} --- with generating responses to questions posted on Stack Exchange~\cite{h4stackexchange}. For images, we used the GenImage dataset \cite{genimage} to compare real ImageNet \cite{imagenet} images to synthetic ones from eight different generative image models. In both cases (Figure~2A,D), we see that the synthetic samples are shifted from the real ones in the initial PCs. Further, Linear Discriminant Analysis (LDA), a simple supervised classification method, finds linear boundaries that separate samples coming from different sources. 


On top of the ability to distinguish content from different models, we find that embeddings of samples produced from related models are more similar than those from very different models, while still being distinguishable by a simple classifier. For example, Stable Diffusion v1.4 and v1.5 share the same architecture and are fine-tuned from the same base parameters, albeit for a different number of steps \cite{genimage}. A LDA classifier confirms that the outputs of these two models have higher similarity with each other, compared to images from other models (Table~\ref{table:gen_image_binary}). As further evidence, we produced synthetic cat images with three image generation models. Images by two of the models that share the same architecture (Stable Diffusion XL v1.0~\cite{podell2024sdxl} and Open-Dalle v1.1~\cite{OpenDalle}) embed nearby each other. Notably, images produced by the Denoising Diffusion Probabilistic model (DDPM)~\cite{ddpm} embed nearby the cat images from its own training corpus, the LSUN dataset. All four samples of cat images are still easily distinguished by LDA (Figure~2B). This separability induces a phylogeny which indicates that DDPM did not learn to emulate the distribution of images in its own training data, nor are the similarities between the two text-to-image models enough to prevent us from distinguishing them. We note that these samples are also easily distinguished from the cat images in the Cats\&Dogs dataset (Table~\ref{tab:lda_accs_cv}).

Keeping a fixed generative model, FEs can detect variation between prompting techniques that are used to generate data. Figure~2C shows the result of taking abstracts of economics papers from arXiv and having LLaMa-2 70B generate synthetic counterparts using two different prompt templates. While samples from the two prompts have similar position in the PCs, we find that both generated samples are easily distinguished from each other and from the human-made set by LDA.

Even with only a few samples of AI-generated data, dimensionality reduction on the vectors produced by FEs can be used to quickly scan for spurious data. We test whether outlier detection tools can be used to filter a pool of real Stack Exchange replies that have been contaminated by a small batch of responses produced by LLaMa-2 70B. The scatter plots in Figure~2E show that while the synthetic responses fit among the real data in the top few PCs, they are revealed as outliers when we consider more principal components. Empirically, we find that our ability to detect fake responses with this method is related to the proportion of fake answers; too many contaminates creates a new cluster, and too few buries their signal (Figure~\ref{fig:isofor-trend}).

We note that the observed separability between real and AI-generated data only becomes clear when the models are given the freedom to reveal their internal biases. A common feature of the results discussed thus far is that the AI-generated data are relatively unconstrained. In those cases where we have access to real baselines, we are consistently able to separate generated data from real data with over $98\%$ accuracy (Table~\ref{tab:lda_accs},~\ref{tab:lda_accs_cv}). As we impose constraints, such as limiting the length of generated text, the likelihood may drop that the embedded geometries can be disentangled. Using language models to translate short-form text samples across six languages curated by the United Nations~\cite{ziemski-etal-2016-united} and comparing to the accompanying human translations, we find that we cannot reliably predict whether a given translation was done by a machine or by a human (Table~\ref{tab:lda_accs}). These samples are typically one sentence or shorter in length (Figure~\ref{fig:short-long}) and offer few opportunities for a model to make mistakes or demonstrate revealing stylistic choices.

\subsection*{From shifted representations to statistical explanations}
Regression analysis with manually designed explanatory variables lets us connect the numerical features identified by the chosen FE with interpretable patterns in our data. We collected pairs of scientific abstracts, where each pair contains one from the arXiv and one generated by LLaMa-2 70B. Figure~3A shows that across five subject areas, the shift in the PC coordinates between the real abstracts and those generated by LLaMa-2 70B can be linked to the language model's use of words that hyperbolize the results. More regressions for the text experiments presented in Figure~2 are in Table~\ref{tab:main_regressions}. In particular, we observed that the shift between real and synthetic Stack Exchange responses in Figure~2A can be linked to explicit attempts by the language models to be overtly helpful. In another case, the shift that is visually observed in Figure~2C between real and AI-generated economics abstracts can be attributed to the tendency of LLaMa-2 70B to stress the broader implications of the work through words like `innovation,' `valuable,' and `insight', in comparison to reference abstracts from the arXiv. Alternatively, this shift can be attributed to the LLM-generated abstracts being considerably longer than the real abstracts on average. These differences indicate the utility of regression analysis in uncovering explanations for shifts in the embedded samples. In our case, the geometric misalignment between the two kinds of embedded data can be partially sourced by this shifted vocabulary or style.

By exploring the features encoded by FEs, we find that embeddings can be used to automatically organize our data. Just as unsupervised learning on vectorized data can automatically cluster real and generated samples, we find that it also reveals metadata-like qualities. Similar to the topical separation among the multilingual news articles in Figure~1C, we find that four of the top five PCs each distinguished one of the five academic subject areas represented in the collection of real and synthetic abstracts across five subject areas (Figure~3B, see Figure~\ref{fig:arxiv5_pc5} for PC5). The resulting clusters contain the abstracts in that subject area, independent of whether it was real or synthetically generated. Further, LDA achieves high accuracy in identifying both topic and origin (Figure~3C).


\subsection*{Learned features are low dimensional}
Our results thus far exhibit the pervasive phenomena that the embedding vectors produced by a FE are strongly influenced by the sampling techniques used to gather the input data. In Figure~\ref{fig:scree-plot}, we show that the signal of data that is embedded by pre-trained FEs occupies a relatively low-dimensional subspace of the learned embedding vector space. This gives a potential explanation as to why these FEs can easily observe distributional shifts between populations: the signal from a collection of sample data threads a small subspace of a much larger vector space, so we can expect that minor differences in sampling techniques can lead to a shift away from this fixed subspace. In the context of generative models and associated gDNA, precisely capturing the patterns of their training data involves learning a low-dimensional signal in the presence of many dimensions of noise. Further, the existence of gDNA suggests that generative models each learn their own distinct representation, as a consequence of different training corpora and training procedures.


\subsection*{Discussion}


Even if a model produces realistic looking examples, its statistical biases are revealed under the lens of a FE. These shifts are likely caused by an accumulation of many differences in the models (e.g. training parameters, architecture, training corpus, etc.), and we reiterate that this is a general phenomenon well-known in other experimental sciences, such as batch effects in biology. 
Additionally, for token-based predictors like LLMs, repeated sampling serves to accentuate these differences when generating large amounts of tokens. 

The shifts we observed are likely inherent to each generative model and therefore cannot be removed without considerable filtering or human intervention.
This is motivated by the apparent low dimension of data, as illustrated through the embeddings (Figure~\ref{fig:scree-plot}). When a generative model learns a representation of its training data in a high-dimensional vector space, any noise encoded in the extraneous dimensions can cause the resulting generated data to drift from the training data. Further, if the data subspaces have dimension well below the ambient space, then the distributions which describe the real and synthetic data will be concentrated in spaces with nearly zero volume. Thus, a generative AI system has very low probability of matching the true data distribution.

One consequence of our ability to identify AI-generated data is that FEs can protect us from AI-generated fraud. The release of ChatGPT quickly led to noticeable shifts in writing styles~\cite{geng2024chatgpt}, with other studies issuing concern on the human ability to detect articles produced by generative AI models~\cite{Gao2022.12.23.521610, knott2023generative, nightingale2022ai}. This work provides a reasonable foundation for automated validation using embeddings from a FE, potentially giving a quick filter against recent waves of artificially generated scientific papers ~\cite{gray2024chatgpt, jain2023generative}. 
Additionally, the embedded vector representations can give powerful insights to interpret the patterns and biases within the observed data. When applied to AI-generated data, these insights can be used for model selection or prompt tuning through comparative analysis of FE vectorizations. Current techniques for this may not be entirely reliable, as they depend on expensive benchmarks and instance-based evaluation which often requires an LLM judge that is prone to hallucinations and other biases~\cite{kim2024prometheus, manakul-etal-2023-selfcheckgpt, rafailov2023direct}. Directly comparing representations via a FE can let the practitioner study the robustness of generative models. As AI models quickly advance in performance, we expect there will be value in the ability to readily measure for preferential features.



It may be that generative models are not optimized to produce entirely realistic data. 
For natural language, the training process for LLMs is likely geared towards generating helpful and safe content through Reinforcement Learning from Human Feedback~\cite{xu2021recipessafetyopendomainchatbots}, rather than creating human-like text. While these techniques show promise, the misalignment between real and simulated data brings into question what these models are optimizing for, if not to learn the true semantics encoded in language and images. For instance, it is hypothesized that generative models are learning to deceive humans rather than learning ground truths \cite{wen2024languagemodelslearnmislead}. This is further underscored by recent results showing that data produced by generative models are \emph{necessarily} prone to hallucination~\cite{kalai2024calibrated} and that repeated training on such data can lead to model collapse~\cite{shumailov2024nature}. With the internet being flooded with AI-generated content, accessible tools to diagnose our models can help us combat these possibilities.

\newpage


\begin{figure}[ht]
    \centering
    \includegraphics[width=0.95\linewidth]{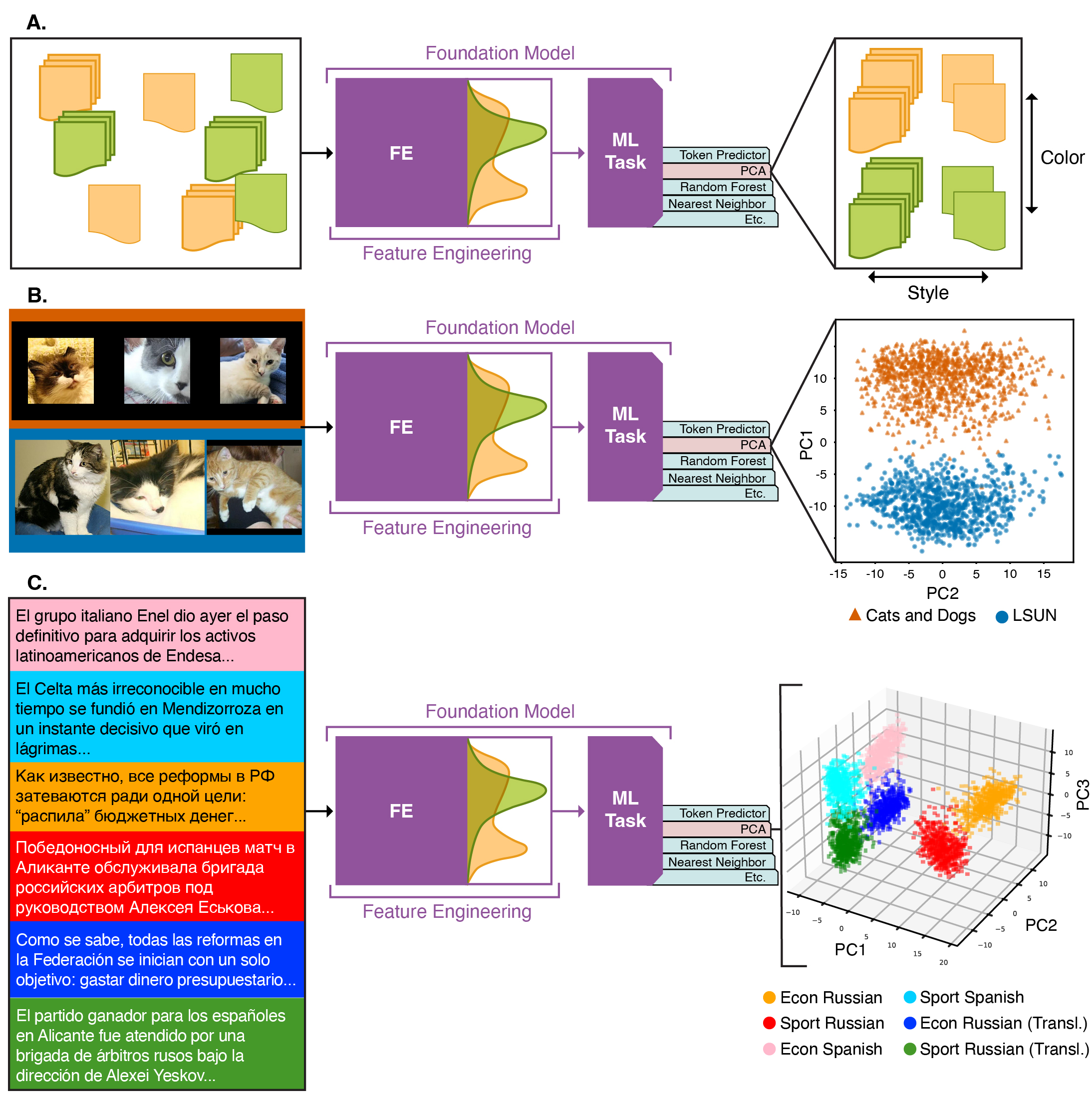}
    \caption{(\textbf{A}) Framework of our techniques. The full neural network begins by engineering a vector representation of the data through its feature embedder (FE). The embedded data is then passed through an interchangeable machine learning layer, such as Principal Component Analysis (PCA). (\textbf{B}) Sample images of cat images from the LSUN dataset and Cats\&Dogs dataset, embedded by Apple's Data Filtering Network. (\textbf{C}) Top three principal components (PCs) on Spanish and Russian news articles, including Russian-to-Spanish machine translations, data embedded by Microsoft's multilingual-e5-large. In order, PCs correspond to language, topic, and natural or translated text.
 }
    \label{fig:draft_fig1}
\end{figure}


\begin{figure} 
	\centering
        \includegraphics[width=0.95\textwidth]{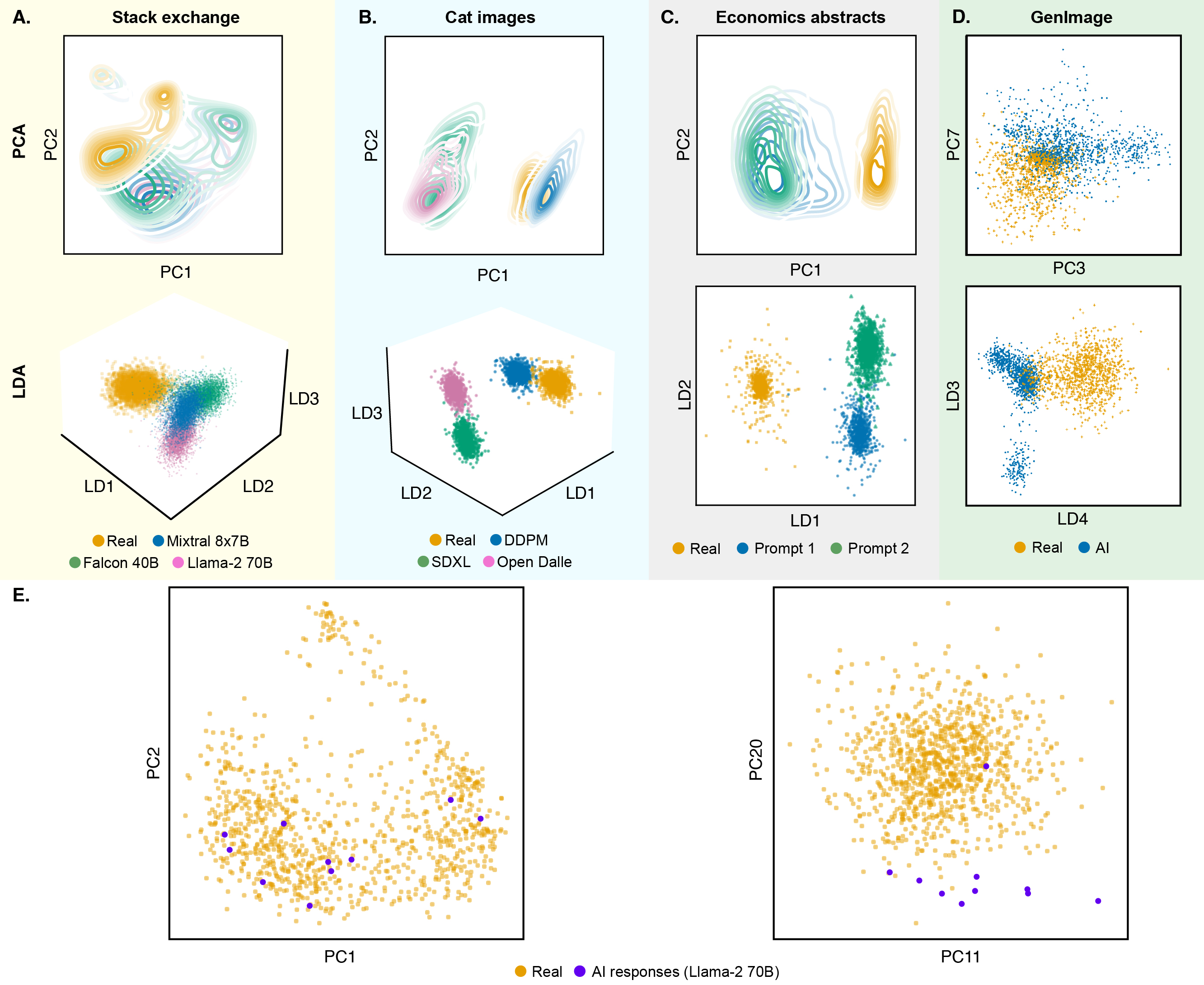}

	\caption{
    \textbf{Embedded data to distinguish real and fake data.} DNN embeddings of AI-generated data alongside real data, after dimensionality reductions. (\textbf{A})
        User-posted Stack Exchange answers and responses by three language models (Mistral 8x7B, Llama-2 70B, and Falcon 40B). Embeddings by Mistral 7B. (\textbf{B}) Real cat images from the LSUN dataset and images produced by three image generation models (DDPM, Stable Diffusion, and Open-Dalle). Embeddings by Apple’s Data Filtering Network. (\textbf{C}) Real economics abstracts and two synthetic samples created by Llama-2 70B using different prompting methods. Embeddings by Mistral 7B. (\textbf{D}) Real and AI-generated images from the GenImage dataset containing data produced by eight generative models and ImageNet. Embeddings by Apple’s Data Filtering network.  (\textbf{E}) Principal components of embedded Stack Exchange responses, contaminated with AI-generated answers by Llama-2 70B. Embeddings by Mistral 7B. AI-generated answers are revealed as outliers when projecting to PCs 11 and 20. 
}
	\label{fig:MAIN2} 
\end{figure}

\begin{table} 
\centering
\begin{tabular}{@{}lllllll@{}}
\hline
 & Response & Explanatory & $R^2$ & $r$ & $F$-stat. \\
\hline
Stack Exchange   & PC 1 $-$ PC 2 & synthetic   & 0.36 & 0.60 & $2.6\times10^4$ \\
(Fig 2A)         & PC 1 & special char ratio & 0.55 & 0.74 & $5.8\times10^4$ \\
                 & LD 1 & phrases* & 0.29 & 0.53 & $5.0\times10^3$ \\
Econ Abstracts   & PC 1 & Real/AI & 0.82 & 0.90 & $1.1 \times 10^4$ \\ 
(Fig 2C)         & PC 1 & length $< 1500$ chars. & 0.57 & 0.76 & $3.4\times10^{3}$ \\
                 & PC 1 & word appearance** & 0.54 & -0.74 &  $3.0\times10^{3}$ \\
\hline
\end{tabular}
\caption{\textbf{Regression on PCA and LDA clusters.} Results of linear regression on principal
component and linear discriminant clusters in various experiments. The listed explanatory variable
represents an indicator independent variable corresponding to the stated attributed of the observed
data, such as whether certain words appear in the text sample. In all cases listed the $p$-values for the $F-$statistics are $0.0$. Further experiments are in Table~\ref{tab:regressions}. \\ *At least two of the following: `alternatively', `example', `helps', `if you have any questions', `worth mentioning', `additionally', `note', `in this case', `apologize', `you are correct', `ultimately', `this shows', `in conclusion', `:\textbackslash n', `AI language model', and appearance of lists. \\ **At least one occurrence of any of `innovation', `valuable', `insight', `demonstrates', `understanding`, `implication'.}
\label{tab:main_regressions}
\end{table}

\begin{figure}[ht]
    \centering
    \includegraphics[width=1.0\textwidth]{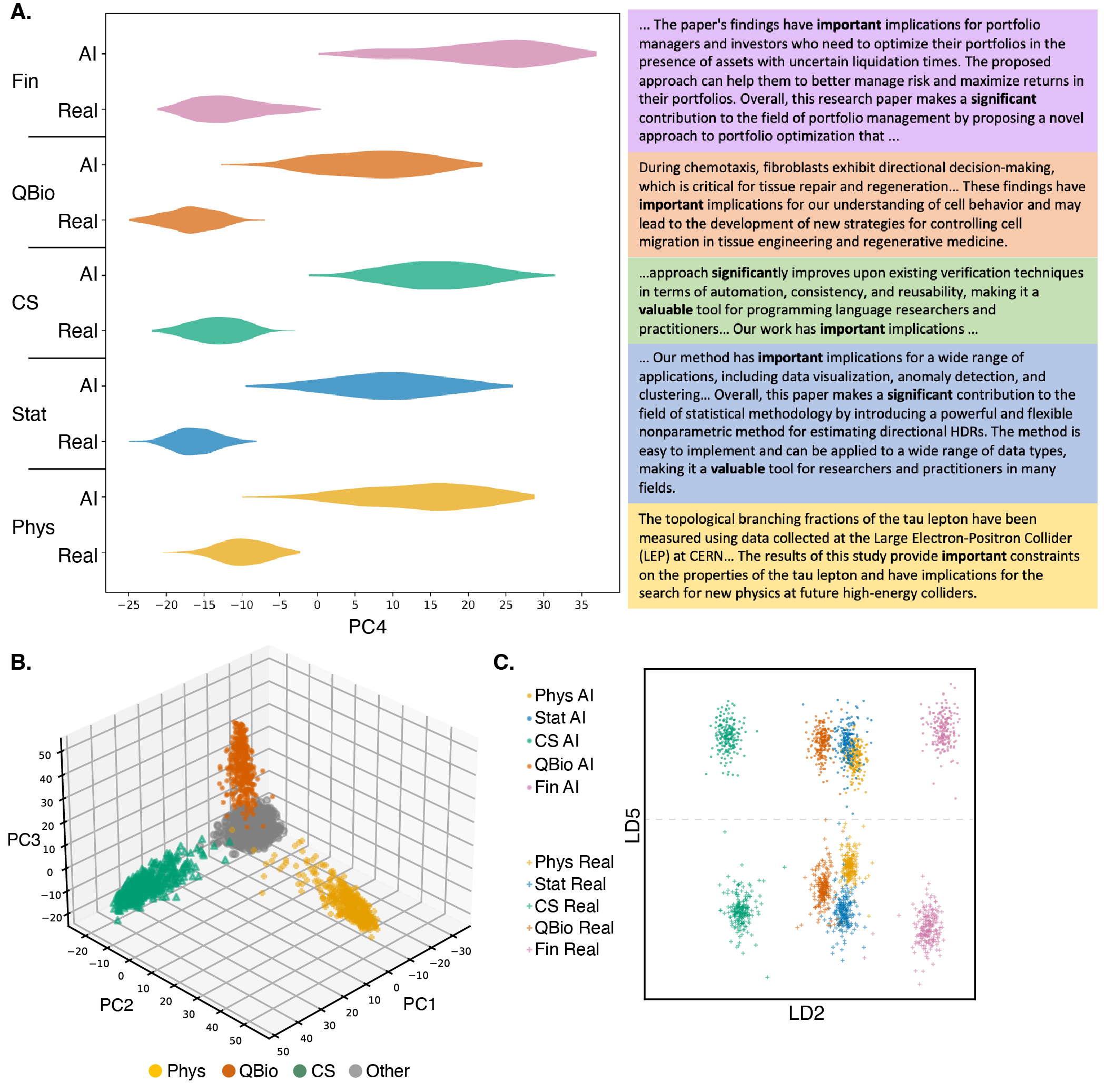}
    \caption{\textbf{Organizing data through embeddings.} \textbf{(A)} We can deduce explanations for the differences in real and AI-generated abstracts using regressions on the PCs; we find the high occurrence of the words 'important,' 'significant,' or 'valuable' highly correlate with the shift in PC4 (see Table~\ref{tab:regressions} for details). \textbf{(B)} Top three principal components on embedded abstracts yield clustering of physics, biology, and computer science subjects. \textbf{(C)} Supervised learning through LDA easily distinguished all ten abstract sources: five subject areas by real and AI. Test acc: $99.0\% \pm 0.47\%$. }
    \label{fig:draft_fig3}
\end{figure}





\clearpage 
\bibliography{main} 
\bibliographystyle{sciencemag}


\section*{Acknowledgments}

The authors thank Tim Marrinan (PNNL), Sinjini Banerjee (Rutgers University), and Michael Robinson (American University) for helpful discussions.


\paragraph*{Funding:}

The authors were partially supported by the Statistical Inference Generates kNowledge for Artificial Learners (SIGNAL) program at PNNL. This work was also supported, in part, by the Office of Defense Nuclear Nonproliferation Research and Development within the U.S. Department of Energy’s National Nuclear Security Administration and Pacific Northwest National Laboratory, which is operated by Battelle Memorial Institute for the U.S. Department of Energy under contract DE-AC05-76RL01830.



\paragraph*{Author contributions:}
TC and ADS conceived the original idea for this work. TC,
MV, 
RC,
AE,
ADS all contributed and  developed the methodological framework while MV and RC performed the experiments. Everyone assisted in the preparation of the manuscript.

\paragraph*{Competing interests:} MV, RC, AE, and ADS do not have competing interests to declare. TC is currently employed by Strategic Analysis Incorporated and is contracted as a Senior Advisor at the Advanced Research Projects Agency for Health (ARPA-H) within the Department of Health and Human Services of the Executive Branch of the US Government.

\newpage


\renewcommand{\thefigure}{S\arabic{figure}}
\renewcommand{\thetable}{S\arabic{table}}
\renewcommand{\theequation}{S\arabic{equation}}
\renewcommand{\thepage}{S\arabic{page}}
\setcounter{figure}{0}
\setcounter{table}{0}
\setcounter{equation}{0}
\setcounter{page}{1} 

\newpage


\subsection*{Materials and Methods}\label{app:methods}

\subsubsection*{Methods Overview}

We view the initial layers of a DNN as a feature embedder (\FE) that maps inputs into a high-dimensional latent vector space $\mathbb{R}^n$. Given a data sample $x$ as an input, the DNN computes an embedding vector $y := \FE(x) \in \mathbb{R}^n$. A data set $X$ of $N$ samples maps to a data matrix $Y \in \mathbb{R}^{N \times n}$. In some cases, such as when the DNN is a LLM, the FE consists of all but the final layer and the embedding vectors are obtained from the final hidden layer of the DNN. In other cases, such as when the DNN is an image feature extractor, the output of the DNN is already a high-dimensional vector and the FE \emph{is} the DNN. When beneficial, we refer to $X$ and $Y$ as the \emph{observed} and \emph{embedded} data respectively. In our experiments, we note that the observed data may originate from another DNN. For example, we can take the outputs of generative language models and analyze them using the FE of a different language model.

We use basic techniques from statistics on the data matrix $Y$ under the assumption that the rows of $Y$ are independent and identically distributed vectors. This amounts to the assumption that the data points of $X$ are themselves sampled independently from an underlying distribution. Reasoning about $Y$ allows us to make deductions about the underlying distribution for $X$.  We then rely on Principal Component Analysis (PCA)~\cite{pearson-pca} for unsupervised learning and Linear Discriminant Analysis (LDA)~\cite{fisher-lda} for supervised learning. 

In order to extract embeddings from FEs, we follow instructions provided by the model developers when available. This is the case for many models on the MTEB leaderboard, for instance. In other cases, (for instance Mistral-7B), we extract embeddings by averaging the embedded tensor in the final layer to produce a 1-dimensional vector for each input.
 
We analyze the resulting embedding vectors using the \textsc{scikit-learn} package, with the provided functionality of \texttt{StandardScaler()} to normalize, \texttt{PCA()} to perform PCA, \texttt{LDA()} to perform LDA, and \texttt{IsolationForest()} for outlier detection. 
We use scree plots to determine how many principal components to pass through the application of isolation forests and to examine the dominant signal in embedded data.

\noindent\textbf{Principal Component Analysis} (PCA)~\cite{pearson-pca} is an unsupervised learning algorithm that, given a data points $x_1,\ldots, x_n\in \mathbb{R}^n$, finds linear coordinate directions in $\mathbb{R}^n$ that maximize the variance observed in the data. The principal components turn out to be eigenvectors of the data's covariance matrix and are linear combinations of the original data features so that PC1 explains the most variation, PC2 explains the second most variation, and so on. In order to compute PCA, we use python's ~\textsc{scikit-learn} package~\cite{scikit-learn}, first centering and normalizing with the \texttt{StandardScaler()} object and computing PCA with its \texttt{PCA()} object. 

We use scree plots to determine which principal component features encode important signal within the data and which ones only encode noise. In each of the examples below, the Scree plots possess an `elbow.' The left side of this elbow indicates principal component vectors that encode notable features in our data, while the right of the elbow indicates those that may not.

\textbf{Linear Discriminant Analysis}~\cite{fisher-lda} is a linear classification technique that, given data $x_1,\ldots, x_n\in\mathbb{R}^n$, finds a linear combination of features in $\mathbb{R}^n$ that separate the different classes within the data. In particular, LDA can be thought of as another linear dimensionality-reduction technique (akin to PCA) which maximizes the separation between the class means. We use the implementation of LDA in the ~\textsc{scikit-learn} package.

\textbf{Cluster Regression} is performed so that we can attribute easily explainable features to the different clustering patterns that we observe in PCA and LDA. In short, we use simple and interpretable features to predict the value along a given feature direction. In cases that our embedded data clusters along this direction, this amounts to predicting the cluster of a sample from its more interpretable feature. 

To perform cluster regression on the textual data, we build indicator functions $I_{F}$ for a particular feature $F$ in the oberserved (text, image) data. Typically, $F$ is inferred from metadata or other easily extractable information. Some examples include the subject of academic abstracts, or whether texts contain certain words or phrases. Using these indicator functions, we produce a linear regression on the clusters observed along principal components and linear discriminants and report the $R^2$ value and the result of an F-test.

\textbf{Isolation Forests}~\cite{isolation-forest} is a technique to identify outliers, or anomalies, within a dataset. It assumes that outliers are fewer in count and have feature values much different than inliers, making them easier to separate. Using a tree structure on the feature space, this technique estimates how easily a single point can be isolated; outliers can be isolated with shallow trees and inliers need deep trees. We use the implementation of isolation forests in the Scikit-Learn package.

Fixing the sample size of real data ($N$) and AI contaminants ($M$), we perform PCA and use scree plots to determine the amount of signal to pass through the algorithm. A parameter search across 50 seeds to find the optimal number of trees to use in our isolation forests, up to using 200 trees. We do this for all $N\in\{100,200,300,400,500,1000,2000\}$ and $M\in\{1,2,4,6,8,10,12,15,20\}$.



\subsubsection*{Models}

\noindent\textbf{Feature Embedding Model Details} Before discussing the datasets, we briefly review the embedding models that we used in our main experiments. In all cases, recall that the associated FE comprises the bulk of the DNN and the resulting embedding is obtained as the output of the final layer in the FE. 

\begin{itemize}
\item Mistral-7B~\cite{jiang2023mistral7b} is a large language model with a transformer-based architecture with vocabulary of 32k tokens, has $32$ layers, and uses $13$ billion parameters per token. It has a context window of 8,192 tokens. The FE consists of all layers except for the final token prediction head. That is, embeddings are obtained from the final hidden layer of dimension 4,096.

\item multilingual-e5-large~\cite{wang2024multilingual} is a 24 layer neural networks that generates $1024$ dimensional embedding vectors from text. The model has 560M parameters and supports 100 languages. It has a context window of 512 tokens and long text is truncated to fit within this window. This model is trained to produce embeddings, so the associated FE is the entire network.

\item Data Filtering Network~\cite{data-filtering-networks} is a CLIP model trained on 5B images that were filtered from an uncurated dataset of image-text pairs. It has 1B parameters and can be used to encode both text and images. 
\end{itemize} 

We tested additional models for our classification tasks in Figure \ref{table:LDA_Embeddings_Comparison}. The main text uses the above models, unless otherwise stated, but it is worth noting that each feature embedder generates its own unique embeddings for a provided dataset. Interestingly in the case of NLP, dedicated embedding models that score well on the MTEB leaderboard~\cite{muennighoff2022mteb} were unable to match the performance of Mistral-7B's out-of-box embeddings for our classification tasks.

\bigskip

\noindent\textbf{Generative Model Details}
Here we provide brief descriptions of the generative models we used to create synthetic data. 

\begin{itemize}
\item Llama-2 70B~\cite{touvron2023llama2openfoundation} is a large language model based on the transformer architecture. It has a context window of $4k$ tokens and $80$ layers. It was trained across $2$ trillion tokens. We use this model to generate synthetic scientific abstracts and responses to Stack Exchange queries.

\item Mixtral-8x7B~\cite{jiang2023mistral7b} is a large language model which uses a ``sparse mixture of experts'' framework. It uses a transformer based architecture with inputs of 32k tokens, has $32$ layers, and uses $13$ billion parameters per token. We used this model for generating responses to Stack Exchange questions.

\item Falcon 40B~\cite{almazrouei2023falconseriesopenlanguage} is another large language model based on the transformer architecture. It has $60$ layers with a $2k$ token-length context window. It was trained on $1$ trillion tokens. This model is used to generate synthetic responses to Stack Exchange queries.

\item nllb-200-distilled-600M and nllb-200-3.3B are language models specifically tailored for language translation tasks and is built off of a modified transformer architecture~\cite{fan2020englishcentricmultilingualmachinetranslation}. The models are only trained to handle $512$ tokens at a time, so we perform translations one sentence at a time. For the MLSUM dataset, we use python's nltk package~\cite{bird2009natural} to parse the data into individual sentences.

\item The \text{opus-mt}~\cite{tiedemann-thottingal-2020-opus} series of language models are dedicated translation models. The architecture of these models is based off the transformer. The models can handle $512$ input tokens, so we perform translations one sentence at a time. For the MLSUM dataset, we use python's nltk package~\cite{bird2009natural} to parse the data into individual sentences.

\item Denoising Diffusion Probabilistic Model (DDPM)~\cite{ddpm} is a seminal diffusion-based image generation model with a U-Net~\cite{unet} style architecture. The specific model used here unconditionally generates images of cats and is trained on the cat class of the LSUN dataset. It has 114M parameters.

\item Stable Diffusion XL v1.0 (SDXL)~\cite{podell2024sdxl} is another diffusion-based model. It is a text-conditional model, so the user must provide a prompt to guide the result of the generated images. It again uses a U-Net style architecture and has 2.7B parameters. 

\item OpenDalle~\cite{OpenDalle} shares the same architecture as SDXL.
\end{itemize} 

\subsubsection*{Experiments Setup and Data}

\noindent\textbf{Stack Exchange} This data set, curated by Huggingface, contains questions and responses from the Stack Exchange forum~\cite{h4stackexchange}. Each question comes with a high and low rated user response. Our reference sample consists of high-rated responses, filtering out anything with an external URL. We then take the corresponding questions and generate synthetic responses from three language models: Llama-2 70B~\cite{touvron2023llama2openfoundation}, Falcon 40B~\cite{almazrouei2023falconseriesopenlanguage}, and Mixtral-8x7B~\cite{jiang2024mixtralexperts}. With both the reference and synthetic responses, we see if the FE from Mistral-7B~\cite{jiang2023mistral7b} could distinguish differences between real and synthetic responses and between responses of different synthetic models.

To generate the data, take a random sample of 20,000 questions from the Stack Exchange preference dataset curated by Huggingface~\cite{h4stackexchange}. We require that these questions do not link to external webpages. Wrapping each question around a prompt template and feeding this into the language models produces a set of synthetic responses for each question. We filter out all responses to any question where some LLM did not complete a response due to context length limitations. That is, we assert that the token length of the question and response remains under 8188 tokens for Mixtral-8x7B, under 4092 tokens for Llama-2-70B, and under 2045 tokens for and Falcon 40B, where we use a small buffer to ensure we do not surpass the context length of the language model. We remove any prefixing whitespace to all responses. Noticing that Falcon 40B produces code blocks with CSS tags \texttt{<code>} and \texttt{<\textbackslash code>}, we replaced this with triple ticks \texttt{```} as was observed in the other sources. Lastly, if any question has a response (whether user-generated or synthetic) that includes a URL, we throw away all responses to that question. What remains is four sets of responses to 11,704 questions for a total of 46,816 responses. 
We use this data to test whether feature embedders can separate user-written responses from those of large language models. The use of multiple models allows us to draw conclusions about generative DNA --- that differences in the training data, regularization techniques, learning schedulers, and other factors surface themselves as detectable differences in model outputs.

\bigskip

\noindent\textbf{MLSUM} This dataset contains online news articles across five languages~\cite{scialom2020mlsum}. In the main text, we selected Spanish and  Russian articles in the topics of sports and economics, sampling 500 articles from each (Language, Topic) pair. We wanted to see if the FEs from the multilingual-e5-large~\cite{wang2024multilingual} model could be used to distinguish between language and/or topic. We then use the machine translator nllb-200-distilled-600M~\cite{nllb2022} to map the Russian data into Spanish to see if the FEs would separate natural from translated articles, giving a total of 6 classes with 500 data points each (sports and economics for Russian, Spanish, and translated Russian). This dataset is used to determine if feature embedders can distinguish the differences between languages, news categories, as well as native and non-native text. 

We also randomly sampled 5,000 German sports articles and had three language models (Mixtral-8x7B, nllb-200-3.3B, and opus-mt-de-en) translate them to English. This second dataset is used for the model DNA tests where we see if feature embedders can separate the outputs of different generative models.

To generate the translations between languages using Mixtral-8x7B, we use the following prompt structure and wrap it in a chat template: ``Translate the following text from $<$start\_language$>$ to $<$end\_language$>$. $<$start\_language$>$ Text: $<$text$>$''. We feed the resulting list of tokens into the model and clean up any preliminary text such as `Translation:' in the model response. Translations produced by the nllb series used a translation pipeline object in the Huggingface Transformers package, setting `src\_lang` and `tgt\_lang` to be the start and ending languages, respectively. Translations with the opus model were created by calling \texttt{model.generate()} on the tokenized text. For both the nllb and opus series used pipelines with the Transformers package, proceeding one sentence at a time due to context window limitations of these models. 

\bigskip

\noindent\textbf{arXiv Five Topics} We randomly sampled $200$ arXiv abstracts from each of the following five subject domains: high-energy particle physics (hep-ex), programming languages (cs.PL), quantitative biology --- cell behavior (q-bio.CB), statistical methodology (stat.ME), and quantitative finance --- portfolio management (q-fin.PM). All abstracts corresponded to papers listed exclusively under their category. We also enforced that the corresponding papers were uploaded in the year 2020 or earlier to ensure that they preceded popular AI-tools such as ChatGPT.

To create the synthetic abstracts, we used Llama-2 70B to generate, for each real arXiV paper, a synthetic abstract for a paper with the same title. We use the following prompt: ``Write the abstract of a research paper titled ``$<$Title$>$" in the field of $<$Field$>$." We used the full field name associated with that topic, not the arXiv identifier (for example, "high-energy particle physics" instead of "hep-ex").

To process the data, we remove attached in-text classifiers like `Keywords:' and those used by the Journal of Economic Literature. We removed newline symbols, web addresses, and text stating where code is available. We also removed any prefix and suffix from synthetic abstracts, such as `Sure! Here is the abstract you wanted:'.  We replaced LaTeX symbols formatting with plain text equivalent. We also replaced Greek letters with their English counterpart (e.g. $\tau$ changed to tau). Any otherwise unusable abstracts were removed as well, such as when the paper was retracted by the author and when the model produced nonsense. We kept ``normal" special character that are plainly visible on a QWERTY keyboard, with the exception of the dollar sign, which LaTeX uses for math equations. 

\bigskip

\noindent\textbf{arXiv Economics} We further collected $816$ abstracts from general economics (econ.GN), where each abstract could be cross-listed with any other categories. Using two different prompt templates, we generated synthetic abstracts using the paper title and category using~Llama-2-70B.
The first prompt used was ``Write the abstract of a research paper titled $<$Title$>$ in the fields of $<$Field1$>$ and $<$Field2$>$ and ... $<$FieldN$>$." The second was ``You are a researcher in the fields of $<$Field1$>$ and $<$Field2$>$ and ... $<$FieldN$>$. You and your research team have written a research paper titled $<$Title$>$ Write the abstract for this paper. Describe the research accomplishments concisely and completely, and take care to describe how this paper fits into the big picture of your fields." Both prompts were modified for grammar if the article was listed in only one field.

In addition to the cleaning performed with the arXiv Five Topic samples, we performed an additional filtration step. We used neither the real or synthetic abstracts when the model failed to generate one. For example, the model refused to generate an abstract with the title ``How to Increase Global Wealth Inequality for Fun and Profit,'' so we removed the attempted output along with the real abstract from our data. When the model yielded two versions of an abstract, a ``base'' and ``revised'' version, we deleted the base version and used the revised. 

While the previous dataset arXiv dataset is used to test if feature embedders can separate real and synthetic data, this one also tests if feature embedders can separate data according to the prompting techniques that are used. 

\bigskip

\noindent\textbf{Cat Images} We sampled $900$ cat images from each of the following sources:
\begin{itemize}
    \item Cats\&Dogs dataset \cite{catsvsdogs}.
    \item LSUN Cat dataset~\cite{yu2015lsun}.
    \item Denoising Diffusion Probabilistic Model (DDPM) \cite{ddpm} is a model trained exclusively on the cat images from the LSUN Cat dataset. We used the hugginface version \verb|ddpm-ema-cat-256|.
    \item Stable Diffusion XL v1.0 (SDXL)~\cite{podell2024sdxl} is another diffusion-based model. It is a text-conditional model, so the user must provide a prompt to guide the result of the generated images. It again uses a U-Net style architecture and has 2.7B parameters. 
    \item OpenDalle~\cite{OpenDalle} shares the same architecture as SDXL.
\end{itemize}

We sampled images from the LSUN Cats set by first shuffling the data and then taking the first $900$ from the shuffled set. In this case, the dataset was too large to shuffle entirely; we instead took the first $400,000$ samples (about $24\%$ of the whole), shuffled these, and then took the first $900$ as our sample. For Cats\&Dogs, this was done by filtering out the dog images, shuffling the full cat image partition, and taking the first $900$ images from the shuffled set. 

The LSUN Cats sample were saved uniformly as $256\times256$ pixels, while the synthetic images and those from Cats\&Dogs were saved in their original size.

All images were preprocessed using TensorFlow \cite{tensorflow}. Images were resized to the maximum input size of the embedding model by center-cropping if the images where too large or by padding with zeros around the perimeter of the image if the image was too small. Each image dimension (RGB) was rescaled to $[0,1]$ and then normalized to mean $(0.48145466, 0.4578275, 0.40821073)$ and standard deviation $(0.26862954, 0.26130258, 0.27577711)$ before being passed to the model. 


To generate synthetic images, the DDPM has an unconditional method to create such an image, so we simply call that function. Generating images with the SDXL and OpenDalle models requires a prompt. For this, we crowd-sourced $105$ prompts from within our organization to curate a list of prompts people would use to generate an image of a cat. From here we increased the number of unique prompts according to the following procedure: (1) duplicate each prompt according to the number of words in it, up to a maximum of $13$ and (2) randomly mask words in each of the new prompts. Each word was kept independently with probability $0.6$. We did not mask any of the following words: ``cat," ``cats," ``Cat," ``Cats," ``kitten," ``kitty," ``kittens", ``Kitten", ``Kittens." 


\bigskip

\noindent\textbf{GenImage} The GenImage \cite{genimage} dataset is a large image-based dataset created for the task of classifying AI-generated and human-made images. The real images are taken from ImageNet \cite{imagenet} and the synthetic images were generated using the ImageNet class labels and eight generative models: ADM \cite{dhariwal2021diffusion}, BigGAN \cite{brock2018large}, Glide \cite{nichol2021glide}, Midjourney \cite{midjourney}, Stable DIffusion 1.4 (SDv4) and 1.5 (SDv5)\cite{rombach2022high}, VQDM \cite{gu2022vector}, and Wukong \cite{wukong}. The dataset contains real and synthetic images from 1000 ImageNet categories.

In our experiment we randomly sampled $100$ ImageNet categories. For each of the sampled classes, we sampled $162$ images from each source (8 generative models plus ImageNet). This is the total number of images from each generative model available in the dataset, with the exception of SDv5, for which there were $166$ images per class\cite{genimage}. In case of SDv5 we shuffled the images corresponding to that class before taking the first $162$ images. With ImageNet, we again shuffled the images corresponding to the class before taking the first $162$ samples. We resized all images to the maximum input size with bilinear interpolation without anti-aliasing. This was all done using TensorFlow \cite{tensorflow}. 

\bigskip

\noindent\textbf{UNPC} The United Nations Parallel Corpus is a machine-translation dataset comprising of official records and documents of the United Nations under public domain~\cite{ziemski-etal-2016-united}. It contains human translations in several languages of the provided records. We used machine translators (opus-mt and nllb-3.3B) to translate the test data (4,000 samples) across the given languages. Using feature embedders, we wanted to test whether the embedding space can be used to distinguish real human translations from the synthetic machine translations.



\begin{figure}
\begin{tabular}{@{}lll@{}}
\hline
Experiment & Embedding Model & All classes ($\%$) \\ 
\hline
arXiv Econ & \textbf{Mistral-7B-Instruct-v0.2}~\cite{jiang2023mistral7b} & $94.6 \pm 0.96$ \\ 
& Nomic-embed-text-v1~\cite{nussbaum2024nomic} & $86.1 \pm 1.53$ \\  
arXiv Five Topic & \textbf{Mistral-7B-Instruct-v0.2}~\cite{jiang2023mistral7b} & $99.0 \pm 0.47$ \\ 
& Nomic-embed-text-v1~\cite{nussbaum2024nomic}& $94.9 \pm 1.07$ \\
Cat Images & \textbf{apple/DFN5B}~\cite{data-filtering-networks} & $98.0 \pm 0.37$ \\
& google/vit-large-patch16-224-in21k~\cite{dosovitskiy2020image} & $87.0 \pm 0.97$ \\
& laion/CLIP-ViT-H-14-laion2B-s32B-b79K~\cite{schuhmann2022laionb, Radford2021LearningTV} & $95.4 \pm 0.61$ \\
& openai/clip-vit-large-patch14-336~\cite{Radford2021LearningTV} & $96.2 \pm 0.60$ \\
GenImage & \textbf{Apple/DFN5B}~\cite{data-filtering-networks} & $96.1 \pm 0.18$ \\ 
MLSUM Transl. (DE$\to$EN) & \textbf{mistralai/Mistral-7B-v0.1}\cite{jiang2023mistral7b} & $93.2\pm0.06$ \\ 
& nomic-ai/nomic-embed-text-v1~\cite{nussbaum2024nomic} & $81.4\pm0.07$ \\
UNPC Transl. (FR$\to$EN) & \textbf{mistralai/Mistral-7B-v0.1}\cite{jiang2023mistral7b} & $37.1\pm0.78$ \\
Stack Exchange & \textbf{mistralai/Mistral-7B-v0.1}\cite{jiang2023mistral7b} & $90.7\pm0.26$ \\ 
& \textbf{mistralai/Mistral-7B-Instruct-v0.2}~\cite{jiang2023mistral7b} & $90.8\pm0.24$ \\
& GritLM/GritLM-7B~\cite{muennighoff2024generative} & $83.8\pm0.43$ \\
& nomic-ai/nomic-embed-text-v1~\cite{nussbaum2024nomic} & $69.9\pm0.44$ \\
& intfloat/e5-large-v2~\cite{wang2022text} & $68.4\pm0.43$ \\
\hline
\end{tabular}
\caption{
\textbf{Embedding models comparison.} Comparison of test accuracy across all classes of the listed experiment of LDA applied to embeddings from respective models is provided. Boldface indicates that that model was used for the results presented in this work, unless otherwise indicated.}
\label{table:LDA_Embeddings_Comparison}
\end{figure}

\begin{figure}
    \centering
    \includegraphics[width=0.49\linewidth]{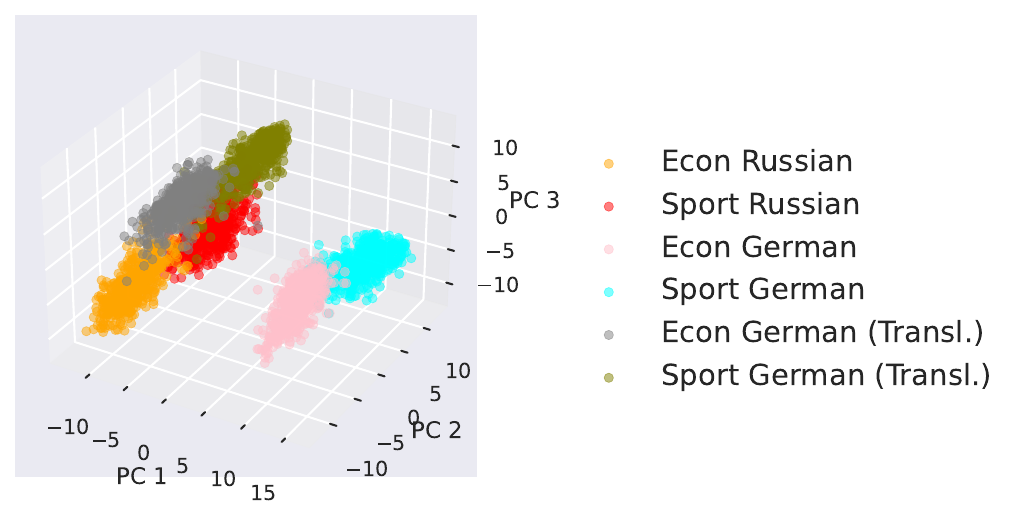}
    \includegraphics[width=0.49\linewidth]{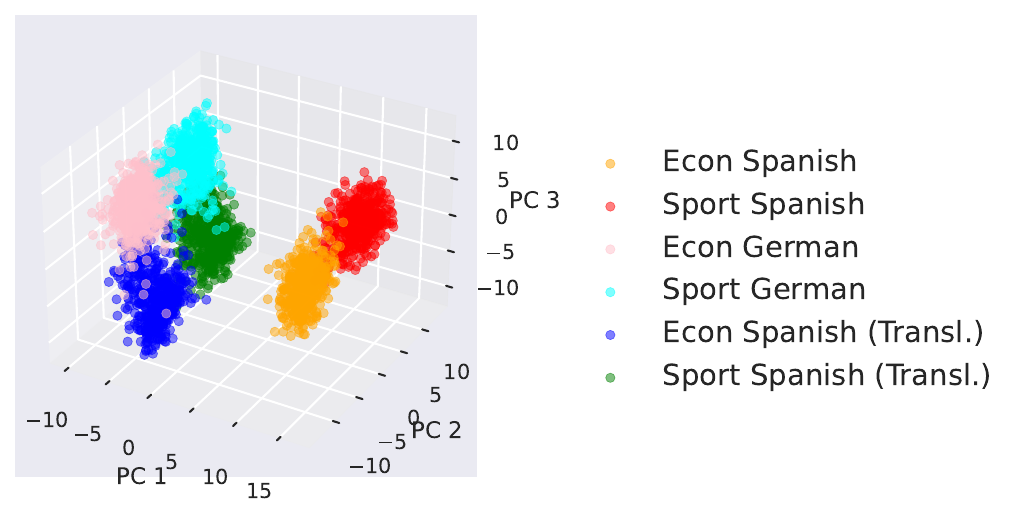}\\
    \includegraphics[width=0.49\linewidth]{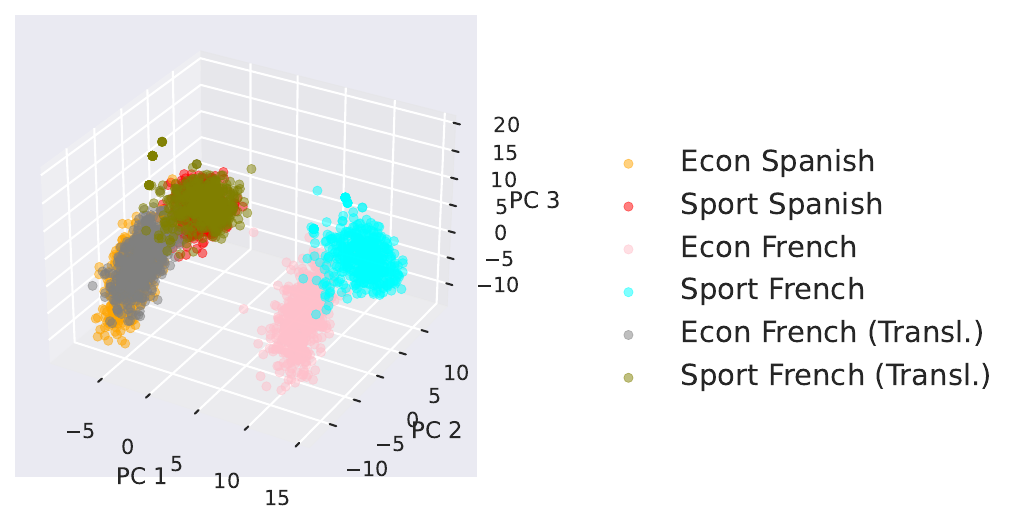}
    \includegraphics[width=0.49\linewidth]{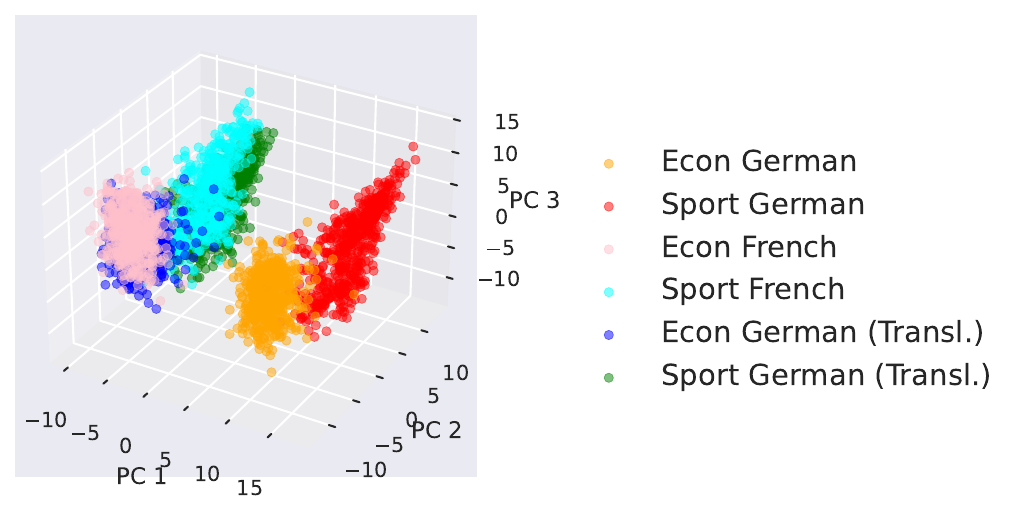}\\
    \caption{Top three components of multilingual news data for various language pairs.}
    \label{fig:other-langs}
\end{figure}

\begin{figure}\label{nmf-topic}
    \centering
    \includegraphics[width=1.0\linewidth]{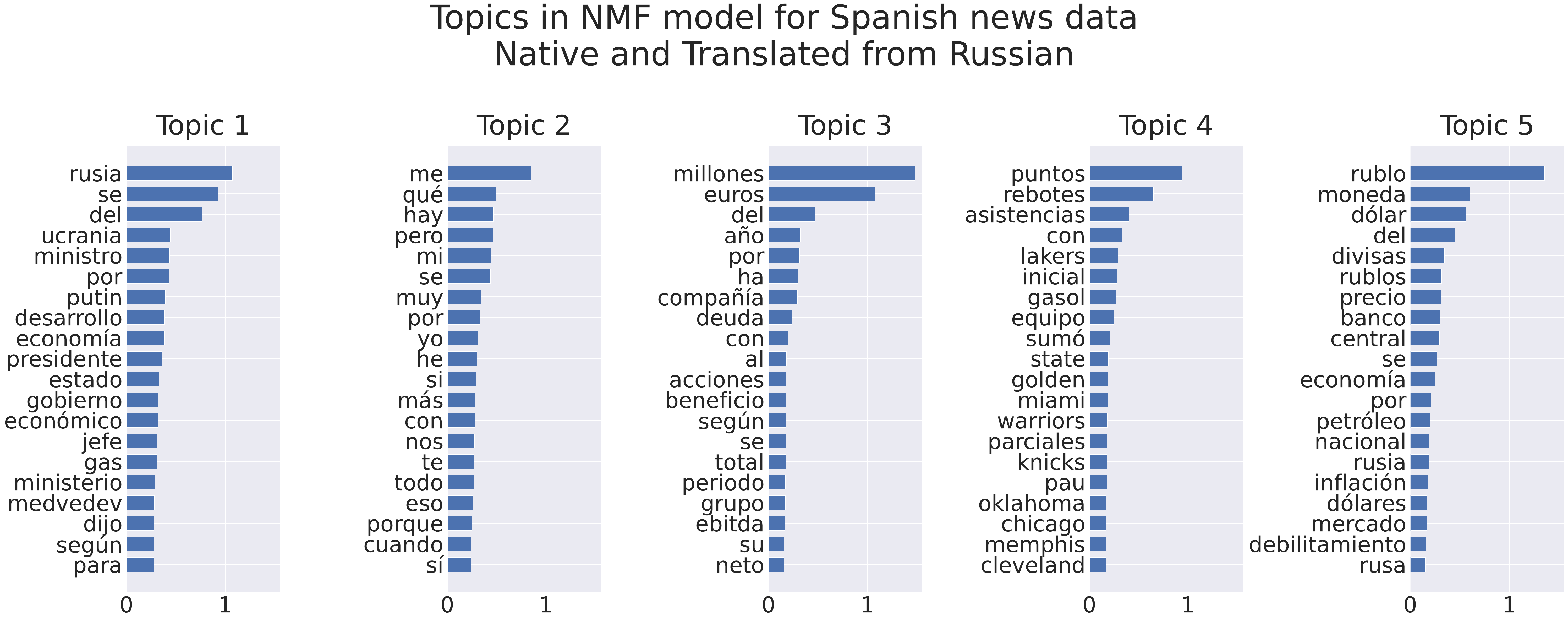}
    \caption{Word Clusters for Top 5 Topics found by NMF}
    \label{fig:nmf_mlsum_factors}
\end{figure}

\begin{figure} 
\centering
\begin{tabular}{@{}lllllll@{}}
\hline
Experiment & Classes & Avg. Test Acc. (\%) & SD (\%)\\
\hline
Stack Exchange   & Real, Llama2 70B & 99.7 & 0.35\\
                 & Real, Mixtral 8x7B & 99.0 & 0.75 \\
                 & Real, Falcon 40B & 98.2 & 0.98 \\
                 & Llama2 70B, Mixtral 8x7B & 87.1 & 3.05 \\
                 & Llama2 70B, Falcon 40B & 98.0 & 1.02 \\
                 & Mixtral 8x7B, Falcon 40B & 94.0 & 1.71 \\
UNPC (ES$\to$EN) & Real, NLLB 3B & 56.7 & 1.11 \\
                 & Real, Opus & 53.6 & 1.06 \\
                 & NLLB 3B, Opus & 51.4 & 1.21 \\
UNPC (FR$\to$EN) & Real, NLLB 3B & 55.4 & 0.83 \\
                 & Real, Opus & 54.0 & 1.19 \\
                 & NLLB 3B, Opus & 51.4 & 1.04 \\
MLSUM Sports     & NLLB 3B, Opus & 94.5 & 0.48\\
(DE$\to$EN)      & NLLB 3B, Mixtral 8x7B & 93.9 & 0.47 \\
                 & Opus, Mixtral 8x7B & 96.4 & 0.40 \\
                 & NLLB 3B, NLLB 600M & 75.7 & 0.87 \\
                 & Mixtral 8x7B, Mistral 7B & 70.2 & 0.94 \\
ArXiv Five Topic & Real, Llama2 70B & $99.9$ & $0.13$ \\
                 & $($Real, Llama2 70B$)$ $\times$ 5 topics & $99.0$ & $0.47$ \\
ArXiv Econ       & Real, Prompt 1, Prompt 2 & $94.6$ & $0.96$ \\
                 & Real, Prompts Combined & $99.8$ & $0.16$ \\
                 & Prompt 1, Prompt 2 & $92.0$ & $1.46$ \\
\hline
\end{tabular}
\caption{\textbf{LDA accuracies across NLP experiments.} LDA accuraces across $50$ train/test splits for different subsets of data in each experiment. For creating embeddings, we use the default (bolded) embedding model corresponding to the respective experiment listed in Figure \ref{table:LDA_Embeddings_Comparison}.} 
\label{tab:lda_accs}
\end{figure}

\begin{figure}[ht]
\begin{tabular}{@{}llllllllll@{}}
\hline
& ADM & BigGAN & Glide & Midjourney & SDV4 & SDV5 & VQDM & Wukong \\
\hline
ADM & -- \\
BigGAN & $100$ & -- \\
Glide & $100$ & $100$  & -- \\
Midjourney & $100$ & $100$& $98.66$ & -- \\ 
SDV4 & $100$ & $100$& $98.17$ & $97.28$ & -- \\ 
SDV5 & $100$ & $100$ & $97.97$ & $96.02$ & $94.85$ & -- \\
VQDM & $100$ & $100$ & $98.26$& $99.00$ & $98.88$ & $98.61$ & --\\
Wukong & $100$ & $100$ & $99.25$ & $98.78$ & $97.60$ & $96.55$ & $96.07$ & -- \\
ImageNet & $100$ & $100$ & $>99.99$ & $99.95$ & $99.99$ & $99.88$ & $99.86$ & $99.75$ \\
\hline
\end{tabular}
\begin{tabular}{@{}llllllllll@{}}
\hline
& ADM & BigGAN & Glide & Midjourney & SDV4 & SDV5 & VQDM & Wukong \\
\hline
ADM & -- \\
BigGAN & $\pm 0.0$ & -- \\
Glide & $\pm 0.0$ & $\pm 0.0$  & -- \\
Midjourney & $\pm 0.0$ & $\pm 0.0$& $\pm 0.20$ & -- \\ 
SDV4 & $\pm 0.0$ & $\pm 0.0$& $\pm 0.29$ & $\pm 0.25$ & -- \\ 
SDV5 & $\pm 0.0$ & $\pm 0.0$ & $\pm 0.34$ & $\pm 0.32$ & $\pm 0.40$ & -- \\
VQDM & $\pm 0.0$ & $\pm 0.0$ & $\pm 0.25$& $\pm 0.21$ & $\pm 0.33$ & $\pm 0.24$ & --\\
Wukong & $\pm 0.0$ & $\pm 0.0$ & $\pm 0.22$ & $\pm 0.23$ & $\pm 0.33$ & $\pm 0.27$ & $ \pm 0.40$ & -- \\
ImageNet & $\pm 0.0$ & $\pm 0.0$ & $\pm 0.01$ & $\pm 0.03$ & $\pm 0.02$ & $\pm 0.06$ & $\pm 0.09$ & $\pm 0.09$ \\
\hline
\end{tabular}
\caption{\textbf{GenImage LDA Binary Classification Results}: Average test accuracy (\%) (top) and standard deviation (\%) (bottom) for binary classifications by LDA of embeddings from sources in GenImage dataset.}
\label{table:gen_image_binary}
\end{figure}

\begin{figure}
    \centering
    \begin{tabular}{@{}lllllll@{}}
\hline
Experiment & Classes & Avg. Test Acc. (\%) & SD (\%)\\
\hline
Cat Images       & LSUN, All Models & $98.8$ & $0.38$ \\
                 & LSUN, DDPM, SDXL, Open-Dalle & $97.5$ & $0.67$\\
                 & Cats \& Dogs, LSUN, DDPM, SDXL, Open-Dalle & $98.0$ & $0.37$ \\
                 & LSUN, DDPM & $92.7$ & $1.42$ \\
                 & SDXL, Open-Dalle & $90.1$ & $1.63$ \\
                 & $($LSUN, DDPM$)$ vs $($SDXL, Open-Dalle$)$ & $100$ & $0.0$ \\
GenImage         & ImageNet vs All 8 Models & $98.3$ & $0.07$ \\
\hline
    \end{tabular}
    \caption{\textbf{LDA accuracies across image experiments.} LDA accuraces across $50$ train/test splits for different subsets of data in each experiment. For creating embeddings, we use the default (bolded) embedding model corresponding to the respective experiment listed in Figure \ref{table:LDA_Embeddings_Comparison}.} 
\label{tab:lda_accs_cv}
\end{figure}

\begin{figure}
    \centering
    \includegraphics[width=0.45\linewidth]{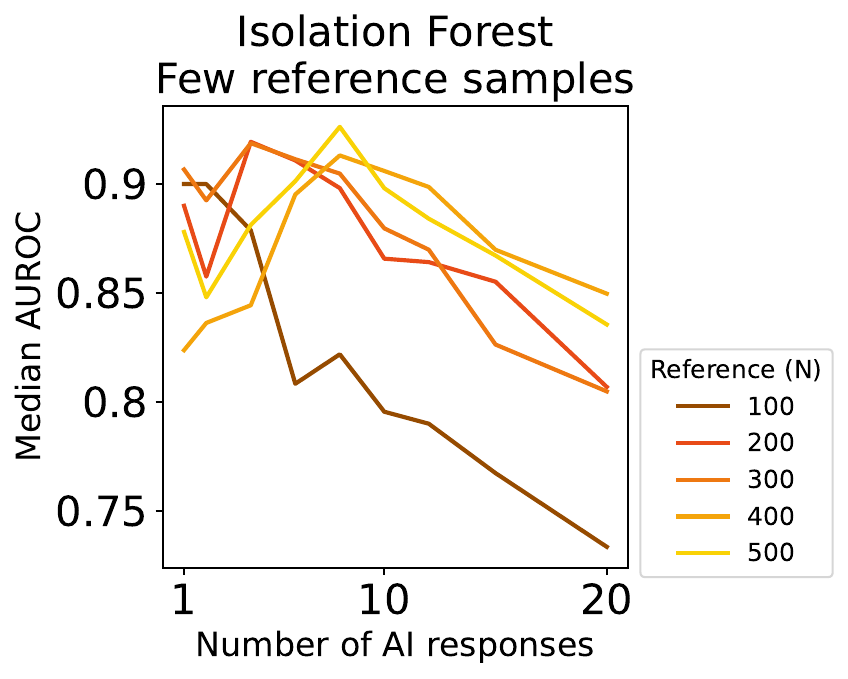}
    \includegraphics[width=0.45\linewidth]{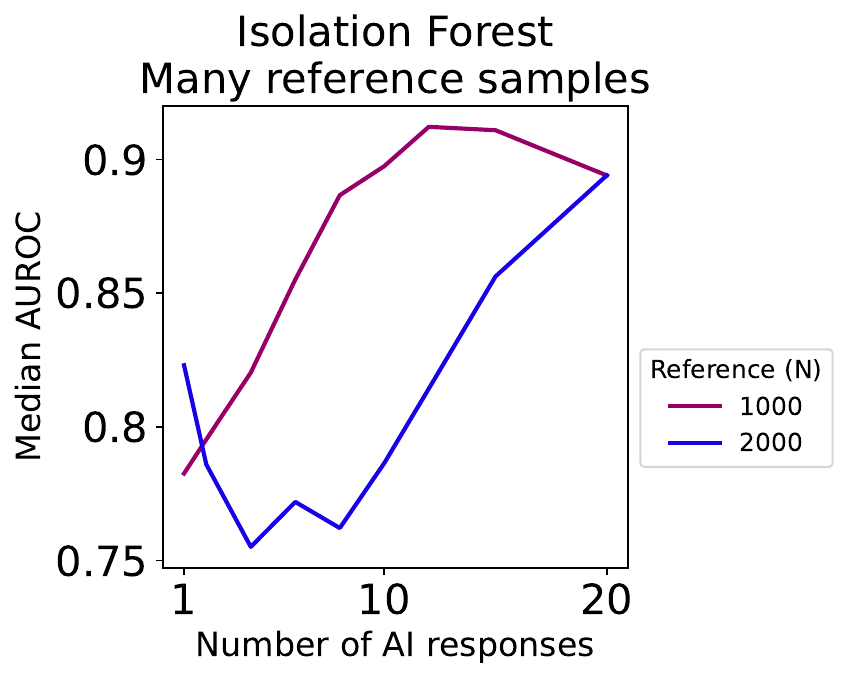}
    \caption{ROC scores for identifying AI-generated answers that contaminate sets of user-posted Stack Exchange questions. Left: With small samples of real answers, increasing the number of AI-responses decreases our ability to identify them as outliers. Right: With large samples of real answers, increasing the number of AI-responses increases our ability to identify them as outliers}
    \label{fig:isofor-trend}
\end{figure}

\begin{figure}
    \centering
    \includegraphics[width=0.7\linewidth]{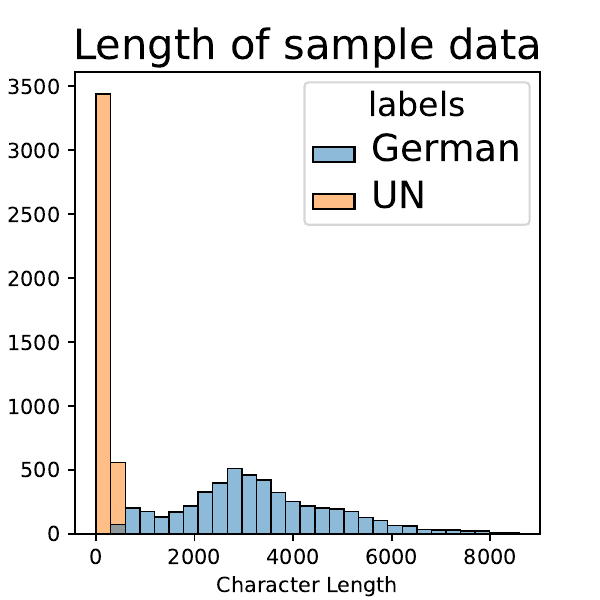}
    \caption{Distribution of news article lengths in the MLSUM dataset along with the lengths of texts in the UNPC testing dataset.}
    \label{fig:short-long}
\end{figure}

\begin{figure}
    \centering
    \includegraphics[width=0.7\linewidth]{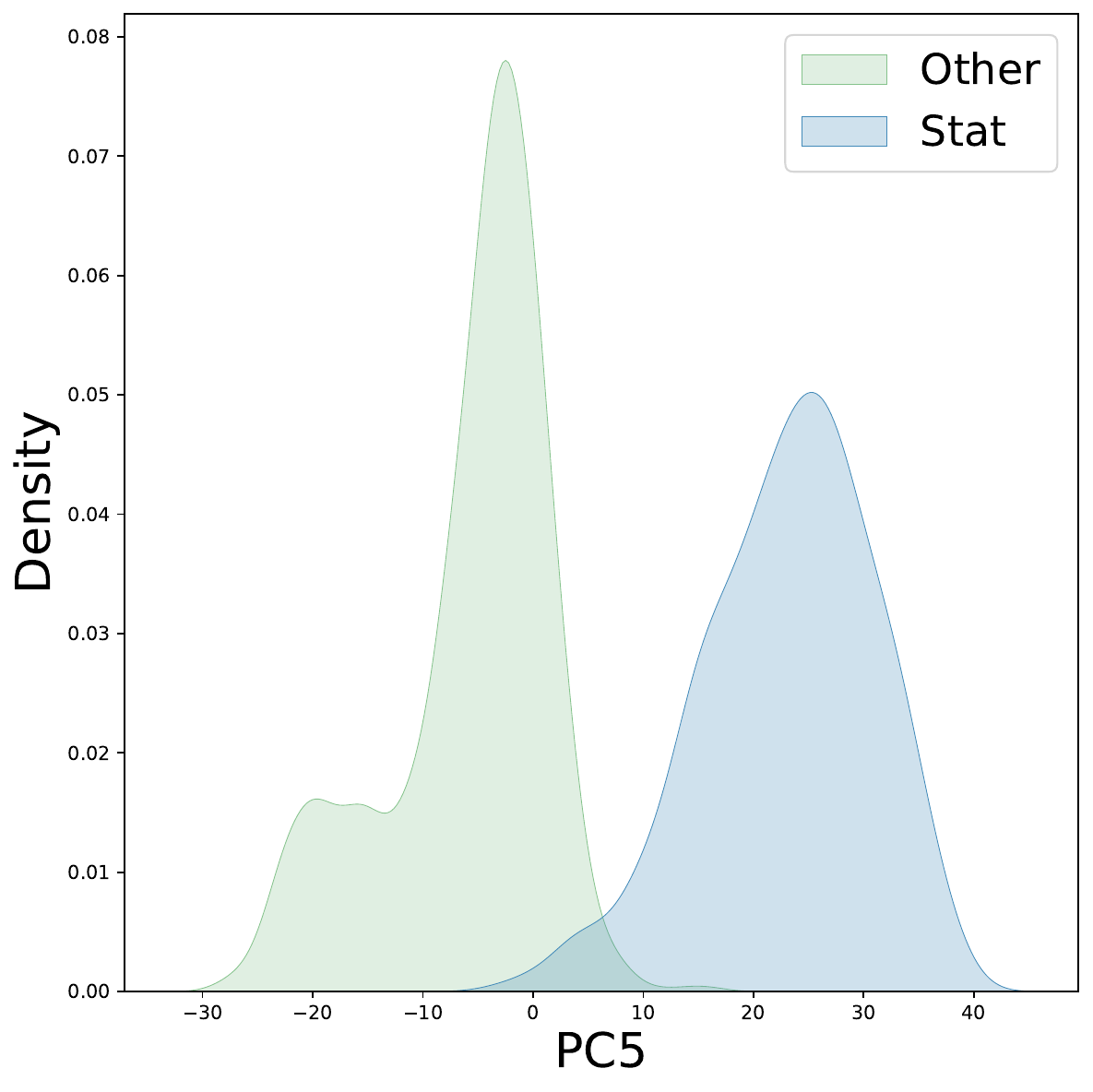}
    \caption{\textbf{PC5 clusters statistics abstracts.} The fifth principal component of the five-topic abstracts clusters the embeddings by whether they were in the statistics subject area, independent of if they originated from the arXiv or the LLaMa-2 70B.}
    \label{fig:arxiv5_pc5}
\end{figure}

\begin{figure}
    \centering
    \includegraphics[width=0.3\linewidth]{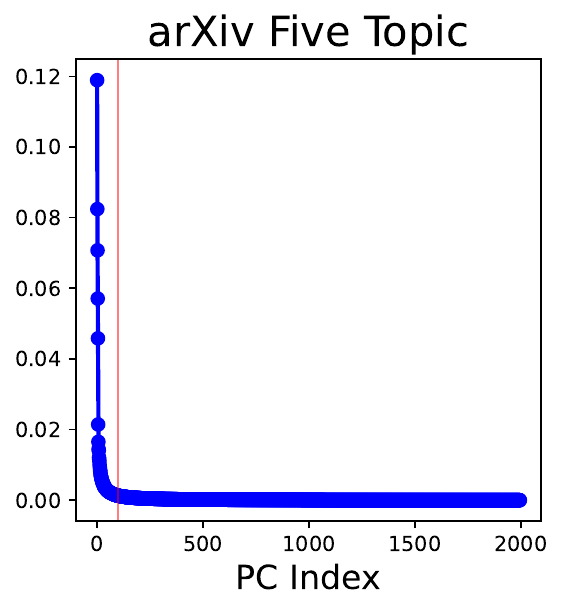}
    \includegraphics[width=0.3\linewidth]{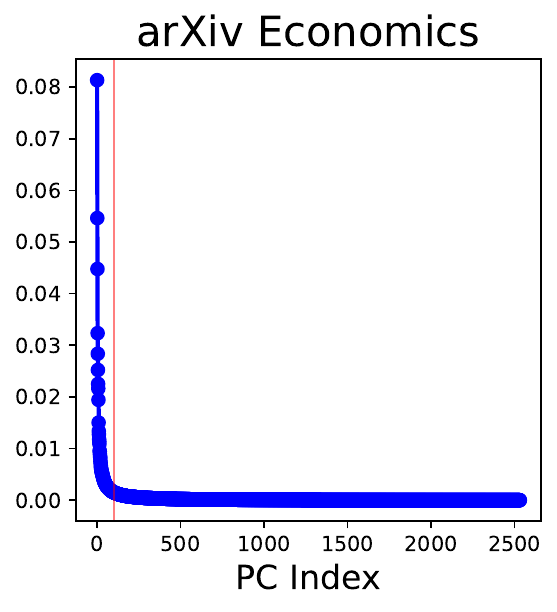}
    \includegraphics[width=0.3\linewidth]{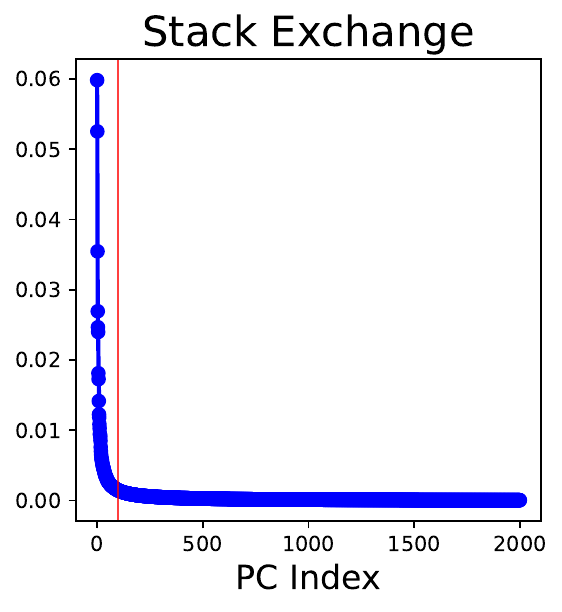}\\
    \includegraphics[width=0.3\linewidth]{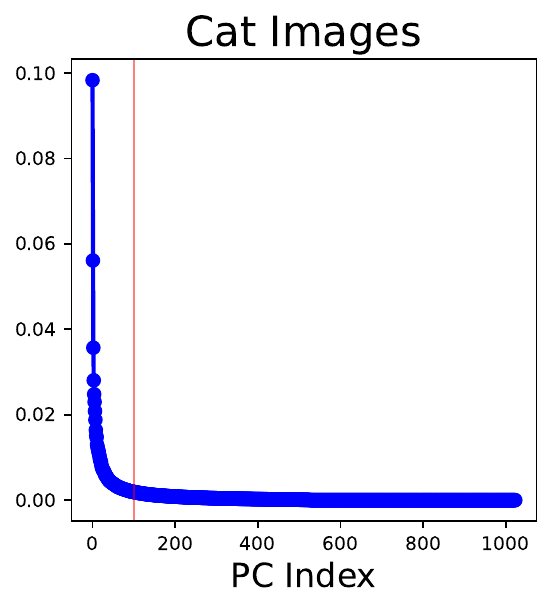}
    \includegraphics[width=0.3\linewidth]{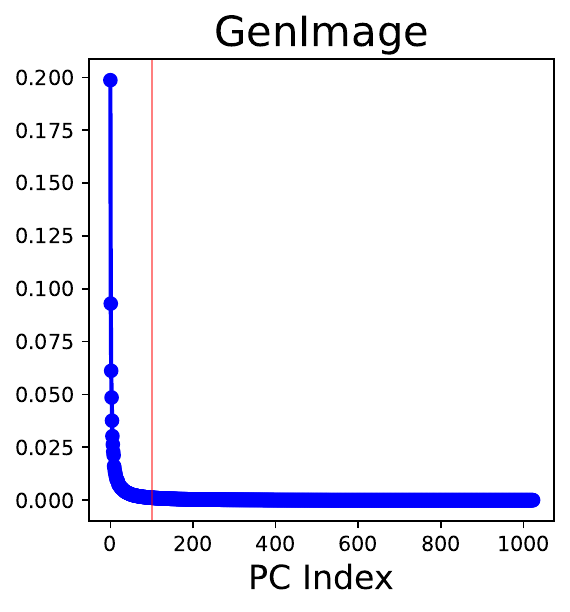}
    \caption{\textbf{Scree plots for various datasets.} For each dataset, the red vertical line indicates the top 100 principal components of the embedded features. \textbf{Top row}: Embeddings by Mistral 7B. Left: Five-topic abstracts from both the arXiv and LLaMa-2 70B. Red line represents $75.3\%$ of cumulative explained variance. Middle: Economics abstracts from both the arXiv and LLaMa-2 70B. Red line represents $70.1\%$ of cumulative explained variance. Right: Stack Exchange responses generated by three language models alongside user-posted responses. Red line represents $60.2\%$ of cumulative explained variance. \textbf{Bottom row}: Embeddings by Apple's Data Filtering Network. Left: Cat images from the LSUN dataset, Cats\&Dogs dataset, SDXL, Open-Dalle, and DDPM. Red line represents $75.8\%$ of cumulative explained variance. Right: Real and syntheitc mages from the GenImage dataset. Red line represents $89.2\%$ of cumulative explained variance.}
    \label{fig:scree-plot}
\end{figure}

\begin{figure} 
\centering
\begin{tabular}{@{}lllllll@{}}
\hline
 & Response & Explanatory & $R^2$ & $r$ & $F$-stat. & $p$ (for $F$) \\
\hline
Translation      & PC 1 & language   & 0.98 & 0.99 & $1.5\times10^5$ & 0.0\\
                 & PC 2 & topic & 0.87 & 0.93 & $2.5\times10^4$ & 0.0\\
                 & PC 3 & natural/translated* & 0.83 & 0.92 & $1.0\times10^4$ & 0.0\\
Stack Exchange   & PC 1-2 & synthetic   & 0.36 & 0.60 & $2.6\times10^4$ &0.0\\
                 & PC 1 & special char ratio & 0.55 & 0.74 & $5.8\times10^4$ & 0.0 \\
                 & PC 1 & lists & 0.17 & 0.42 & $9.8\times10^3$ & 0.0 \\
                 & LD 1 & synthetic & 0.84 & 0.92 & $7.2\times10^4$ &  0.0 \\
                 & LD 1 & phrases** & 0.29 & 0.53 & $5.0\times10^3$ & 0.0 \\
ArXiv Five Topic & PC 1 & Phys   & 0.92 & 0.96 & $2.4\times10^{4}$ & 0.0  \\
                 & PC 2 & CS & 0.76 & 0.87 & $6.3\times10^{3}$ & 0.0  \\
                 & PC 3 & QBio & 0.86 & 0.93 & $1.2\times10^{4}$ & 0.0  \\
                 & PC 4 & Real/AI & 0.79 & 0.89 & $7.3 \times10^{3}$& 0.0  \\
                 & PC 4 & word appearance$\dagger$ & 0.47 & 0.68 & $1.8\times10^{3}$ &      $1.0\times10^{-270}$ \\
                 & PC 5 & Stats.  & 0.71 & 0.84 & $4.9\times10^{3}$ & 0.0  \\
ArXiv Econ       & PC 1 & Real/AI & 0.82 & 0.90 & $1.1 \times 10^4$ & 0.0 \\ 
                 & PC 1   & length $< 1500$ chars. & 0.57 & 0.76 & $3.4\times10^{3}$ &  0.0  \\
                 & PC 1   & word appearance\S & 0.54 & -0.74 &  $3.0\times10^{3}$ &         0.0 \\
                 & LD 2 & word appearance\P & 0.41 & 0.64 & $1.2\times10^{3}$ &             $4.2\times10^{-195}$ \\
LSUN Cats        & PC 1 & Lineage & 0.88 & 0.94 & $2.7 \times 10^4$ & 0.0 \\
\hline
\end{tabular}
\caption{\textbf{Regression analysis on PCA clusters.} Results of linear regression on principal component and linear discriminant clusters in various experiments. The listed explanatory variable  represents an indicator independent variable corresponding to the stated attributed of the observed data, such as whether certain words appear in the text sample. *When restricted to the target language. **At least two of the following: `alternatively', `example', `helps', `if you have any questions', `worth mentioning', `additionally', `note', `in this case', `apologize', `you are correct', `ultimately', `this shows', `in conclusion', `:\textbackslash n', `AI language model', and appearance of lists. The removal of `AI language model' marginally effects the value ($R^2=0.27$ and Pearson $r=0.52$.) $\dagger$ At least one occurance of any of `significant', `important', or `valuable'. \S At least one occurance of any of `innovation', `valuable', `insight', `demonstrates', `understanding`, `implication'. \P More than four occurrences of either `we' or `our'.}
\label{tab:regressions}
\end{figure}

\clearpage 



\end{document}